\relax
\documentclass[letterpaper]{article} 
\usepackage{aaai21}  
\usepackage{times}  
\usepackage{helvet} 
\usepackage{courier}  
\usepackage[hyphens]{url}  
\usepackage{graphicx} 
\usepackage{natbib}  
\usepackage{caption} 
\usepackage{amsmath}
\usepackage{amssymb}
\usepackage{subcaption}
\usepackage[switch]{lineno}  %
\usepackage[dvipsnames]{xcolor}
\usepackage{multirow}

\usepackage[symbol]{footmisc}

\nocopyright
\title{Faster Re-translation Using Non-Autoregressive Model For Simultaneous Neural Machine Translation}
\author{Hyojung Han$^*$, Sathish Indurthi$^*$, Mohd Abbas Zaidi, Nikhil Kumar Lakumarapu, \\ Beomseok Lee, Sangha Kim, Chanwoo Kim, Inchul Hwang \\}

\affiliations{
    Samsung Research, Seoul, South Korea \\
    \{h.j.han, s.indurthi, abbas.zaidi, n07.kumar, bsgunn.lee, sangha01.kim, chanw.com, inc.hwang\}@samsung.com}

\begin{document}
\maketitle

\begin{abstract}
Recently, simultaneous translation has gathered a lot of attention since it enables compelling applications such as subtitle translation for a live event or real-time video-call translation. Some of these translation applications allow editing of partial translation giving rise to re-translation approaches. The current re-translation approaches are based on autoregressive sequence generation models (\textit{ReTA}),  which generate target tokens in the (partial) translation sequentially.
During inference time, the multiple re-translations with sequential generation in \textit{ReTA} models lead to an increased wall-clock time gap between the incoming source input and the corresponding target output as the source input grows.
In this work, we propose a faster re-translation system based on a non-autoregressive sequence generation model (\textit{FReTNA}) to alleviate the huge inference time incurring in \textit{ReTA} based models.
 The experimental results on multiple translation tasks show that the proposed model reduces the average computation time (wall-clock time) by a factor of 10 when compared to the \textit{ReTA} model by incurring a small drop in the translation quality. It also outperforms the streaming based \textit{Wait-k} model both in terms of the computation time (1.5 times lower) and translation quality.
\end{abstract}

\begin{figure*}[t]
 \label{fig:arch}
    \centering
    \includegraphics[width=\textwidth]{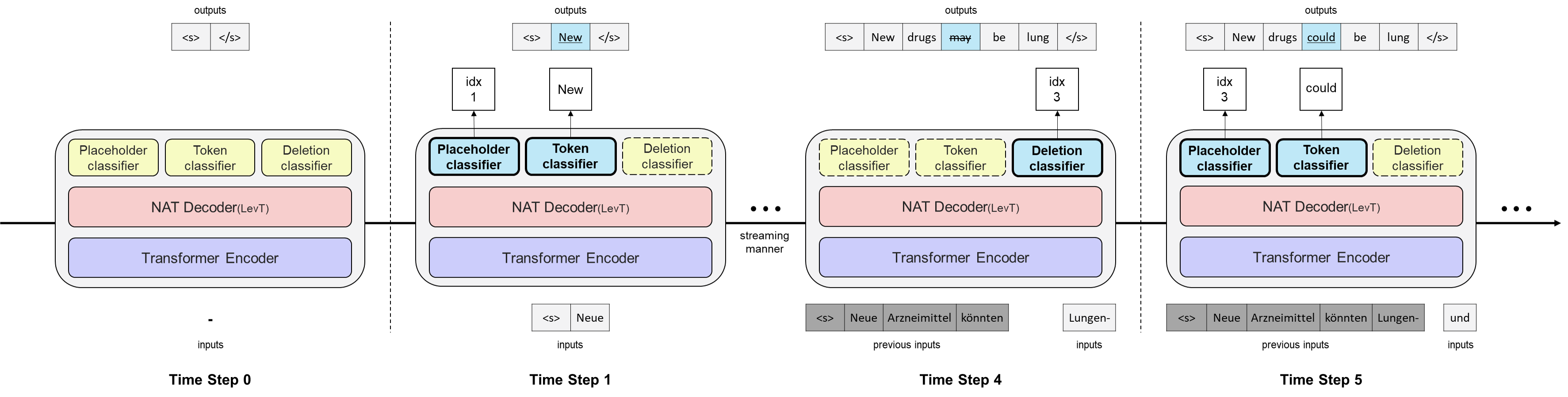}
    \caption{Overview of the proposed \textit{FReTNA} and illustrated using German-to-English example.}
    \label{fig:arch}
\end{figure*}

\footnotetext[1]{Equal contribution}
\section{Introduction}
Simultaneous Neural Machine Translation (SNMT) addresses the problem of real-time interpretation in machine translation. In order to achieve live translation, an SNMT model alternates between reading the source sequence and writing the target sequence using either a fixed or an adaptive policy. Streaming SNMT models can only append tokens to a partial translation as more source tokens are available with no possibility for revising the existing partial translation. A typical application is conversational speech translation, where target tokens must be appended to the existing output.

For certain applications such as live captioning on videos, not revising the existing translation is overly restrictive. Given that we can revise the previous (partial) translation simply re-translating each successive source prefix becomes a viable strategy.
The re-translation based strategy
is not restricted to preserve the previous translation leading to  high translation quality. The current re-translation approaches are based on the autoregressive sequence generation models (\textit{ReTA}), which generate target tokens in the (partial) translation sequentially. As the source sequence grows, the multiple re-translations with sequential generation in \textit{ReTA} models lead to an increased inference time gap causing the translation to be out of sync with the input stream. 
Besides, due to a large number of inference operations involved, the \textit{ReTA} models are not favourable for resource-constrained devices.

In this work, we build a re-translation based simultaneous translation system using non-autoregressive sequence generation models to reduce the computation cost during the inference. The proposed system generates the target tokens in a parallel fashion whenever a new source information arrives; hence, it reduces the number of inference operations required to generate the final translation. To compare the effectiveness of the proposed approach, we implement the  re-translation based autoregressive  (\textit{ReTA}) \cite{arivazhagan2019retranslation} and \textit{Wait-k} \cite{ma2018stacl} models along with our proposed system. Our experimental results reveal that the proposed model achieves significant performance gains over the \textit{ReTA} and \textit{Wait-k} models in terms of computation time while maintaining the property of superior translation quality of re-translation over the streaming based approaches.

Revising the existing output can cause textual instability in re-translation based approaches. The previous approached proposed a stability metric, Normalized Erasure (NE) \cite{arivazhagan2019retranslation}, to capture this instability. However, the NE only considers the first point of difference between pair of translation and fails to quantify the textual instability experienced by the user. In this work, we propose a new stability metric, Normalized Click-N-Edit (NCNE), which better quantifies the textual instabilities by considering the number of insertions/deletions/replacements between a pair of translations.   

The main contributions of our  work are as follows:
\begin{itemize}
    \item We propose re-translation based simultaneous system to reduce the high inference time of current re-translation approaches.
    \item We propose a new stability metric, Normalized Click-N-Edit, which is more sensitive to the flickers in the translation as compared to existing stability metric, Normalized Erasure. 
    \item We conduct several experiments on simultaneous text-to-text translation tasks and establish the efficacy of the proposed approach.
\end{itemize}

\section{Faster Retranslation Model}
\subsection{Preliminaries} We briefly describe the simultaneous translation system to define the problem and set up the notations. The source and the target sequences are represented as $\mathbf{x}=\{x_1, x_2, \cdots, x_S\}$ and $\mathbf{y}=\{y_1, y_2, \cdots, y_T\}$, with $S$ and $T$ being the length of the source and the target sequences. Unlike the offline neural translation models (NMT), the simultaneous neural translation models (SNMT) produce the target sequence concurrently with the growing source sequences. In other words, the probability of predicting the target token at time $t$ depends only on the partial source sequence ($x_1, \cdots, x_{g(t)}$). The probability of predicting the entire target sequence $\mathbf{y}$ is given by:

\begin{equation}
    p_g(\mathbf{y}|\mathbf{x}) = \prod_{t=1}^{T} p(y_t| \mathcal{E}(\mathbf{x}_{\leq g(t)}), \mathcal{D}(\mathbf{y}_{<t})),
\end{equation}
where $\mathcal{E}(.)$ and $\mathcal{D}(.)$ are the encoder and decoder layers of the SNMT model which produce the hidden states for the source and target sequences. The $g(t)$ denotes the number of source tokens processed by the encoder when predicting the token $y_t$, it is bounded by $0 \leq g(t) \leq S$. For a given dataset $D = \{\mathbf{x}_n, \mathbf{y}_n\}_{n=1}^{N}$, the training objective is,

\begin{equation}
    \ell_g = -\sum_{(\mathbf{x}, \mathbf{y}) \in D} \log p_g(\mathbf{y}|\mathbf{x}).
 \end{equation}

In a streaming based SNMT model \cite{ma2018stacl}, whenever new source information arrives, a new target token is generated by computing $p(y_t| \mathcal{E}(\mathbf{x}_{\leq g(t)}), \mathcal{D}(\mathbf{y}_{<t}))$ and is added to the current partial translation. On the other hand, the re-translation based SNMT models compute  $\prod_{i=1}^{t} p(y_i| \mathcal{E}(\mathbf{x}_{\leq g(t)}), \mathcal{D}(\mathbf{y}_{<i}))$ from scratch every time the new source information arrives. 
Let us assume that the source tokens are coming with a stride $s$, i.e., $s$ input tokens arriving at once, and $O(T \times C)$ is the computational time required to generate the sequence, where $C$ is the computational cost of the model for predicting one target token, then the re-translation based system takes $O(\frac{S}{s} \times T \times C)$ due to the repeated computation of the translation.


\begin{table}[b]
\centering
\begin{tabular}{ |c|c|c|c| } 
 \hline
 Label & (Partial) Translations & NE & NCNE \\ \hline
 prev & I live South Korea and  & - & -  \\ 
 current 1 & I live \underline{in} South Korea, and I am  & 3 & 1 \\ 
 current 2 & I live \underline{in} \underline{North} \underline{Carolina} and I am & 3 & 3 \\ \hline
\end{tabular}
\caption{Examples computations of Normalized Erasure and the proposed Normalized Click-N-Edit stability measures on the previous and current (partial) translations.}
\label{table:ne-cne}
\end{table}

\subsection{Faster Re-translation With NAT}
To address the issue of high computation cost faced in the existing re-translation models, we design a new re-translation model based on the non-autoregressive translation (NAT) approach.

\newcommand{\floor}[1]{\left\lfloor #1 \right\rfloor}


The encoder in our model is based on the Transformer encoder \cite{vaswani2017attention} and the decoder is adopted from the Levenshtein Transformer (LevT) \cite{gu2019levenshtein}. We choose the  LevT model as our decoder since it is a  non-autoregressive neural language model and suits the re-translation objective of editing the partial translation. The overview of the proposed system, referred to as \textit{FReTNA}, is illustrated with a German-English example in the Figure \ref{fig:arch}. We describe the main components of LevT and proposed changes to enable the smoother re-translation in the following paragraphs. 

The LevT model parallelly generates all the tokens in the translation and iteratively modifies the translation by using insertion/deletion operations. These operations are achieved by employing Placeholder classifier, Token Classifier, and Deletion Classifier components in the Transformer decoder. The sequence of insertion operations are carried out by using the placeholder and token classifiers where the placeholder classifier is for finding the positions to insert the new tokens and the token classifier is for filling these positions with the actual tokens from the vocabulary $\nu$. The sequence of deletion operations are performed by using the deletion classifier. The inputs to these classifiers come from the Transformer encoder $(T_E)$ and decoder blocks$(T_D)$ and are computed as:

\begin{equation}
\tiny
    \mathbf{e_{0}^{l}}, \cdots, \mathbf{e_{g(t)}^{l}} = \begin{cases}
    \mathbf{E_{x_0}}+\mathbf{P_0}, \cdots, \mathbf{E_{x_{g(t)}}}+\mathbf{P_{g(t)}},& l = 0\\
    \mathrm{T_E}(\mathbf{e_{0}^{l-1}}, \cdots, \mathbf{e_{g(t)}^{l-1}}),              & l = \{1, \cdots, L\}
    \end{cases}
\end{equation}

\begin{equation}
\tiny
    \mathbf{h_{0}^{l}}, \cdots, \mathbf{h_{t}^{l}} = \begin{cases}
    \mathbf{E_{y_0}}+\mathbf{P_0}, \cdots, \mathbf{E_{y_t}}+\mathbf{P_t},& l = 0\\
    \mathrm{T_D}(\{\mathbf{h_{0}^{l-1}, e_{\leq g(t)}^{l-1}}\}, \cdots, \{\mathbf{h_{t}^{l-1}, e_{\leq g(t)}^{l-1}}\}),              & l = \{1, \cdots, L\}
    \end{cases}
\end{equation}
where $E \in \mathbb{R}^{|\nu| \times d}$ and $P \in \mathbb{R}^{N_{\mathrm{max}} \times d}$ are word and position embeddings of a token. 
The decoder outputs from the last Layer $(h_{t}^{L})$ are later passed to the three classifiers to edit the previous translation by performing insertion/deletion operations. These operations are repeated whenever new source information arrives. 


\paragraph{Placeholder classifier:} It predicts the number of tokens to be inserted between every two tokens in the current partial translation. As compared to the LevT's placeholder classifier, we incorporate a positional bias which is given by the second term in the Eq. \ref{eq:pc}. 
As the predicted sequence length grows, the bias becomes stronger, and the model inserts lesser tokens at the start, reducing the flicker. The placeholder classifier with positional bias is given by:

\begin{multline}
\small
\label{eq:pc}
    \pi_{\theta}^{pc}(p|i, \mathbf{x}_{\leq g(t)}, \mathbf{y}_{i}) = \mathrm{softmax}({\color{ForestGreen}\alpha} * \mathbf{h}.\mathbf{B}^T + {\color{ForestGreen}(1-\alpha)\mathbf{q}} ), \\  {\color{ForestGreen}q_k = \frac{\gamma_i}{k+1}}, \\ \quad i = \{1, \cdots, t-1\}, \quad k=\{0, \cdots, K-1\},    
\end{multline}
where $\mathbf{h}=[\mathbf{h}_i^L:\mathbf{h}_{i+1}^L]$, $\mathbf{B} \in \mathbb{R}^{K \times 2d}$, and {\color{ForestGreen}$\gamma_i=\frac{t-i}{t}$}.  Based on the number of $(0 \sim (K-1))$ of tokens predicted by Eq. \ref{eq:pc}, we insert that many placeholders ($<plh>$) at the current position $i$ and it is calculated for all the positions in the (partial) translation of length $t$. Here, $K$ represents the maximum number of insertions between two tokens and  $\alpha$ is a learnable parameter which balances the predictions based on the hidden states and (partial) translation length.

\paragraph{Token Classifier:} The token classifier is similar to LevT's token classifier, it fills in tokens for all the placeholders inserted by the placeholder classifier. This is achieved as follows:
\begin{equation}
\label{eq:tc}
     \pi_{\theta}^{tc}(v|i, x_{\leq g(t)}, y_i) = \mathrm{softmax}(\mathbf{h_i^L}.C^T), \quad \forall y_i =\phi,
\end{equation}
where $C \in \mathbb{R}^{|\nu| \times d}$ and $\phi$ is the placeholder token.

\paragraph{Deletion Classifier:} It scans over the hidden states $(h_{0}^{L}, \cdots, h_{t}^{L})$ (except for the start token and end token) and predicts whether to \textit{keep}(1) or \textit{delete}(0) each token in the (partial) translation. Similar to the placeholder classifier, we also add a positional bias to the deletion classifier to discourage the deletion of initial tokens of the translation as the source sequence grows. The deletion classifier with positional bias is given by:

\begin{multline}
\label{eq:dc}
    \pi_{\theta}^{dc}(d|i, x_{\leq g(t)}, y_i) = \mathrm{softmax}({\color{ForestGreen}\beta}*\mathbf{h}_i^L.\mathbf{A}^T + {\color{ForestGreen}(1-\beta)\gamma_i * \mathbf{l}}), \\
    {\color{ForestGreen}\mathbf{l}=[0, 1]}  , \quad i = \{1, \cdots, t\},
\end{multline}
where $\mathbf{A} \in \mathbb{R}^{2\times d}$, and we always keep the boundary tokens ($<s>, </s>$).

The model with these modified placeholder and deletion classifiers focuses more on appending the partial translation whenever new source information comes in, which results in smoother translation having lower textual instability. Here,  $\beta$ is a learnable parameter.

The insertion and deletion operations are complementary; hence, we combine them in an alternate fashion. In each iteration, first we call the \textit{Placeholder classifier} followed by \textit{Token classifier}, and the \textit{Deletion classifier}. We repeat this process till a certain stopping condition is met, i.e., generated translation is same in consecutive iterations, or MAX iterations are reached. In our experimental results, we found that two iterations of insertion-deletion operations are sufficient while generating the partial translation for the newly arrived source information. To produce the partial translation, the model incurs $2*Z$ cost, where $Z$ is the cost for insertion-deletion operations equals to $C+\epsilon$ since we also have similar decoding layer. The overall time complexity of  our model (\textit{FReTNA}) is $O(\frac{S}{s} \times Z)$ \footnote{The time complexities provided for \textit{ReTA} and \textit{FReTNA} models are for comparison and do not represent the actual computational costs}, since all the target tokens are generated parallelly. The \textit{FReTNA} computational cost is $\sim T$ times less than the \textit{ReTA} model.

\begin{figure*}[h]
  \centering
    \hspace{-1cm}
    \begin{subfigure}[t]{0.45\textwidth}
      \centering
        \includegraphics[height=9.5cm,width=9cm]{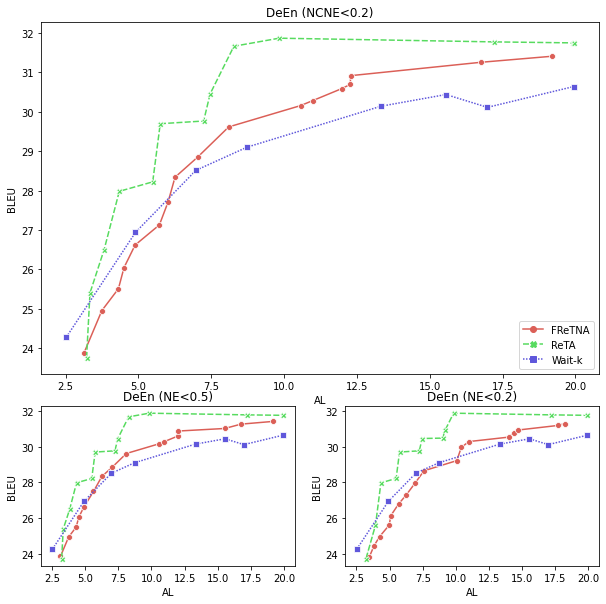}
    \end{subfigure}%
    ~ 
    \hspace{0.8cm}
    \begin{subfigure}[t]{0.45\textwidth}
      \centering
        \includegraphics[height=9.5cm,width=9cm]{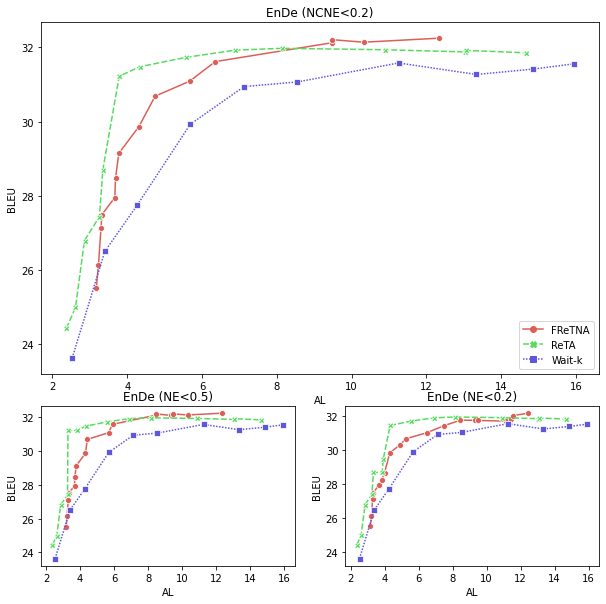}
    \end{subfigure}%
    \caption{
    Quality v/s Latency plots of \textit{FReTNA}, \textit{ReTA}, \textit{Wait-k} models  for the DeEn and EnDe language pairs with different stability constraints.   
    }
    \label{fig:de_ed_bleu}
\end{figure*}

\begin{figure}[t]
    \centering
        \includegraphics[width=9cm, height=9cm]{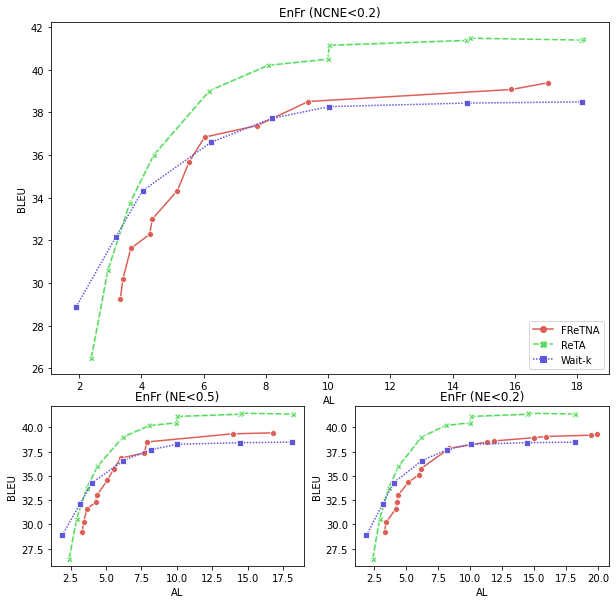}
        
    \caption{%
    Quality v/s Latency plot of \textit{FReTNA}, \textit{ReTA}, \textit{Wait-k} models for EnFr language pair with different stability constraints. 
    }
    \label{fig:en_fr_bleu}
\end{figure}


\begin{figure*}[t]
   \centering
    \hspace{-1cm}
    \begin{subfigure}[t]{0.3\textwidth}
        \centering
        \includegraphics[width=6cm]{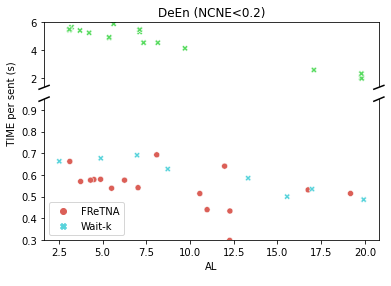}
    \end{subfigure}
    ~ 
    \hspace{0.5cm}
    \begin{subfigure}[t]{0.3\textwidth}
        \centering
        \includegraphics[width=6cm]{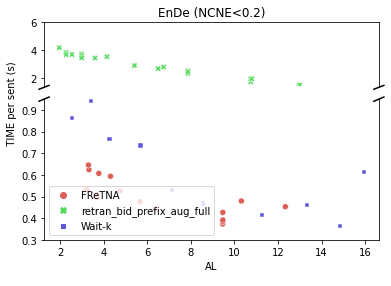}
    \end{subfigure}
    ~
    \hspace{0.5cm}
    \begin{subfigure}[t]{0.3\textwidth}
   
        \centering
        \includegraphics[width=6cm]{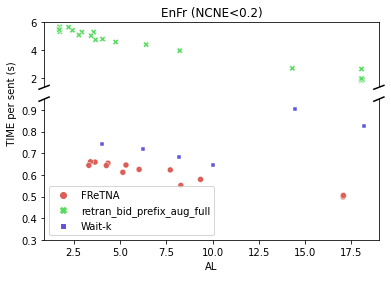}
    \end{subfigure}
    \caption{
    Inference time v/s Latency plot for DeEn, EnDe, and EnFr language pairs of \textit{FReTNA}, \textit{ReTA}, \textit{Wait-k} models for different latency range.
    }
    \label{fig:al_time} 
    
\end{figure*}


\subsection{Training}
\label{sec:lev-training}
We use imitation learning to train the \textit{FReTNA} similar to the Levenshtein Transformer. Unlike \citet{re-translation}, which is trained on prefix sequences along with full-sentence corpus, we train the model on the full sequence corpus only. The expert policy used for imitation learning is derived from a sequence-level knowledge distillation process \cite{knowledge_distillation}. More precisely, we first train an autoregressive model using the same datasets and then replace the original target sequence by the beam-search result of this model. Please refer to \citet{gu2019levenshtein} for more details on imitation learning for LevT model.


    

\subsection{Inference}
\label{sec:inference}
At inference time, we greedily (beam size=1) apply the trained model over the streaming input sequence. For every set of new source tokens, we apply the insertion and deletions policies and pick the actions associated with high probabilities in Eq. \ref{eq:pc}, \ref{eq:tc}, and \ref{eq:dc}. During the re-translation based simultaneous translation, the partial translations are inherently revised when a new set of input token arrives; hence, we apply only two iterations of insertion-deletion sequence on the current partial translation. We also impose a penalty on current partial translation to match the prefix part of the previous translation by subtracting a penalty $\eta$  from the logits in eq. \ref{eq:pc} and eq. \ref{eq:dc}.

\subsection{Measuring Stability of Translation}
\label{sec:stability_metrics}

One important property of the re-translation based models is that they should produce the translation output with as few textual instabilities or flickers as possible; otherwise, the frequent changes in the output can be distracting to the users.
The \textit{ReTA} model \cite{arivazhagan2019retranslation} uses Normalized Erasure (NE) as a stability measure by following \citet{Niehues2016,NiehuesPHSW18,arivazhagan2020retranslation}, it measures the length of the suffix that is to be deleted from the previous partial translation to produce the current translation. However, the metric does not account for the actual number of insertions/deletions/replacements, which provide a much better measure to gauge the visual instability. In the Table \ref{table:ne-cne}, the NE gives same penalty to both the current translations, however, the \textit{current translation 2} would obviously cause more visual instability as compared to the \textit{current translation 1}. In order to have a better metric to represent the flickers during the re-translation, we suggest a new stability measure metric, called Normalized Click-N-Edit (NCNE). The NCNE is computed (Eq \ref{eq:cne}) using the Levenshtein distance \cite{distance_Levenshtein}, which computes the number of insertions/deletions/replacements to be performed on the current translation to match the previous translation. As shown in the Table \ref{table:ne-cne}, the NCNE gives higher penalty to the \textit{current translation 2} since it has a higher visual difference as compared to the \textit{current translation 1}. The NCNE measure aligns better with the textual stability goal of the re-translation based SMT models. The metric is given as

\begin{equation}
\label{eq:cne}
    \mathrm{NCNE} = \frac{1}{T} \sum_{i=2}^{S} \mathrm{levenshtein\_distance}(o_i, o_{i-1}),
 \end{equation}
where $o_i$ and $o_{i-1}$ represent the current and previous translations.

\section{Experiments}
\subsection{Datasets}
We use three diversified MT language pairs to evaluate the proposed model: WMT'15 German-English(DeEn), IWSLT 2020 English-German(EnDe), WMT'14 English-French(EnFr)  data. 

\paragraph{DeEn translation task:}
We use WMT15 German-to-English (4.5 million examples) as the training set. We use \textit{newstest2013} as dev set. All the results have been reported on the \textit{newstest2015}.

\paragraph{EnDe translation task:} For this task, we use the dataset composition given in IWSLT 2020. The training corpus consists of MuST-C, OpenSubtitles2018, and WMT19, with a total of 61 million examples. We choose the best system based on the MuST-C dev set and report the results on the MuST-C \textit{tst-COMMON} test set. The WMT19 dataset further consists of Europarl v9, ParaCrawl v3, Common Crawl, News Commentary v14, Wiki Titles v1 and Document-split Rapid for the German-English language pair. Due to the presence of noise in the OpenSubtitles2018 and ParaCrawl, we only use 10 million randomly sampled examples from these corpora.

\paragraph{EnFr translation task:}
We use WMT14 EnFr (36.3 million examples) as the training set, \textit{newstest2012 + newstest2013} as the dev set, and \textit{newstest2014} as the test set.

More details about the data statistics can be found in the Appendix.

\subsection{Metrics}
\label{sec:metrics}
We adopt the evaluation framework similar to \citet{arivazhagan2019retranslation}, which includes the metrics for quality and latency. The translation quality is measured by calculating the de-tokenized BLEU score using \textit{sacrebleu} script \cite{post-2018-call}. 

Most latency metrics for the simultaneous translation are based on delay vector $g$, which measures how many source tokens were read before outputting the $t^{\text{th} }$ target token. To address the re-translation scenario where target content can change, we use \textit{content delay} similar to \citet{arivazhagan2019retranslation}. The content delay measures the delay with respect to when the token finalizes at a particular position. For example, in the Figure \ref{fig:arch}, the $3^{\text{rd}}$ token appears as \textit{may} at step 4, however, it is finalized at step 5 as \textit{could}. The delay vector $g$ is modified based on this content delay and used in Average Lagging (AL) \cite{ma2018stacl} to compute the latency.

\begin{figure}[t]
 
    \centering
        \includegraphics[width=9cm]{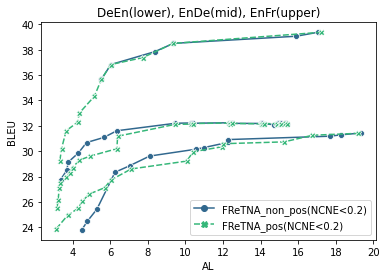}
    \caption{
    Positional bias versus Non-Positional bias. 
    }
    \label{fig:pos-nonpos}
\end{figure}

\subsection{Implementation Details}
The proposed \textit{FReTNA}, \textit{ReTA} \cite{arivazhagan2020retranslation}, and \textit{Wait-k} \cite{ma2018stacl} models are implemented using the \textit{Fairseq} framework \cite{ott2019fairseq}. All the models use Transformer as the base architecture with  settings similar to \citet{re-translation}. The text sequences are processed using word piece vocabulary \cite{subwords}. All the models are trained on 4*NVIDIA P40 GPUs for 300K steps with the batch size of 4096 tokens. The \textit{ReTA} is trained using prefix augmented training data and \textit{FReTNA} uses distilled training dataset as described in \citet{knowledge_distillation}. The hyperparameter $\eta$ described in the Section \ref{sec:inference} is set to $0.2*k$, where $k=\{1, \cdots, K\}$. The Appendix contains more details about the implementation and hyperparameters settings.


\begin{table}[]
\centering

\begin{tabular}{|c|c|c|}
\hline
Pair                       & ReTA                      & FReTNA                    \\ \hline
DeEn                       & 31.7                     & 31.0                     \\ \hline
EnDe                       & 31.8                      & 32.2                     \\ \hline
EnFr                       & 41.2                      & 38.2
     \\ \hline
\end{tabular}

\caption{Performance of offline models on test set of DeEn,EnDe and EnFr pairs with greedy decoding and max iterations set to nine for \textit{FReTNA}.}
\label{table:offline}
\end{table}

\subsection{Results}

In this section, we report the results of our experiments conducted on the DeEn, EnDe and the EnFr language pairs. 
In order to test our \textit{FReTNA} system, we compare it with the recent approaches in re-translation (\textit{ReTA},  \citet{arivazhagan2019retranslation}), and streaming based systems (\textit{Wait-k}, \citet{ma2018stacl}). Unlike traditional translation systems, where the aim is to achieve a higher BLEU score, simultaneous translation is focused on balancing the quality-latency and the time-latency trade-offs. Thus, we compare all the three approaches based on these two trade-offs: (1) Quality v/s Latency and (2) Inference time v/s Latency. The latency is determined by the AL. The inference time signifies the amount of the time taken to compute the output (normalized per sentence). 

\begin{table*}[!h]
\centering

\begin{tabular}{|c|p{7.4cm}|p{7.4cm}|}
\hline
Step & \textit{ReTA} Model & \textit{FReTNA} Model \\ \hline \hline

\multicolumn{3}{|c|}{\textbf{Input Sequence \#1}: Berichten zufolge hofft Indien darüber hinaus auf einen} \\
\multicolumn{3}{|c|}{ Vertrag zur Verteidigungszusammenarbeit zwischen den beiden Nationen.}  \\
\hline
1 & India & Reports \\
2 & India is & India reportedly hopes \\ 
3 & India is also & India is hopes reportedly to hoping for one . \\
4 & India is also re & India is hopes reportedly to hoping for one defense treaty \\ 
5 & India is also reporte & India is reportedly reportedly hoping for one defense treaty between the two nations .\\
6 & India is also reportedly & India is also reportedly hoping for a defense treaty between the two nations .\\ 
7 & India is also reportdly hoping &  -\\
8 & India is also reportedly hoping for & -  \\  
9 & India is also reportedly hoping for a &  - \\
10 & India is also reportedly hoping for a treaty &  - \\
11 & India is also reportedly hoping for a treaty on &  -\\
12 & India is also reportedly hoping for a treaty on defense & - \\ 
13 & India is also reportedly hoping for a treaty on defense cooperation & - \\ 
14 & India is also reportedly hoping for a treaty on defense cooperation between the two nations. & - \\



\hline

\end{tabular}
\caption{Sample translation process by \textit{ReTA} and \textit{FReTNA} models on DeEn pair.}
\label{table:ex}
\end{table*}




\paragraph{Quality v/s Latency:} The Figures \ref{fig:de_ed_bleu} 
and \ref{fig:en_fr_bleu} shows the quality v/s latency trade-off for DeEn, EnDe, and EnFr language pairs.

We report the results on both the NE and NCNE stability metrics  (Section \ref{sec:stability_metrics}). The Re-translation models have similar results with NCNE $< 0.2$ and NE $< 0.5$ metrics. However,  with NE $< 0.2$,  the models have slightly inferior results since it imposes a stricter constraint for stability.   

The proposed \textit{FReTNA} model performance is slightly inferior in the low latency range and better in the medium to high latency range compared to \textit{Wait-k} model for DeEn and EnFr language pairs. For EnDe, our models perform better in all the latency ranges as compared to the \textit{Wait-k} model.


The slight inferior performance of \textit{FReTNA} over \textit{ReTA} is attributed to the complexity of anticipating multiple target tokens simultaneously with limited source context. 
However, \textit{FReTNA}  slightly outperforms \textit{ReTA} from medium to high latency ranges for EnDe language pair. 

\paragraph{Inference Time v/s Latency:} The Figure \ref{fig:al_time} 
shows the inference time v/s latency plots for DeEn, EnDe, and EnFr language pairs.  
Since our model simultaneously generates all target tokens, it has much lower inference time compared to the \textit{ReTA} and \textit{Wait-k} models. 
Generally, the streaming based simultaneous translation models such as \textit{Wait-k} have lower inference time compared to re-translation based approaches such as \textit{ReTA}, since the former models append the (partial) translation whereas the later models sequentially generate the (partial) translation from scratch for every newly arrived source information. Even though our \textit{FReTNA} model is based on re-translation, it has lower inference time compared to the \textit{Wait-k} and \textit{ReTA} models since we adopt a non-autoregressive model to generate all the tokens in the (partial) translation parallelly.


For the comparison purpose, we also trained offline \textit{ReTA}  and  \textit{FReTNA} models for the three language pairs, and the results are reported in Table \ref{table:offline}. The BLEU scores of SMT  and offline models  of \textit{ReTA}  and  \textit{FReTNA}  are comparable. 
Thus, we can conclude that our proposed \textit{FReTNA} approach is better than \textit{ReTA} and \textit{Wait-k} in terms of inference time in all the latency ranges, while maintaining the property of superior translation quality of re-translation over the streaming based approaches.


\paragraph{Impact of positional bias:}
We evaluate the \textit{FReTNA} model with and without including positional bias (FReTNA\_pos vs FReTNA\_non\_pos) introduced in Eq. \ref{eq:pc} and \ref{eq:dc} to see whether positional bias can help the model to generate smoother translations. As shown in Figure \ref{fig:pos-nonpos} FReTNA\_non\_pos has more flickers compared to the FReTNA\_pos model since it's not able to cross the NCNE cutoff of 0.2 in the low latency range. The lower performance of FReTNA\_non\_pos in the low latency range is due to predicting more tokens (insertion policy) than required with less source information. Later, when more source information is available, then some of the tokens have to be deleted (deletion policy), causing more flickers in the final translation output. From Figure \ref{fig:pos-nonpos}, we can see that positional bias reduces flickers in the translation and very useful in low latency range. 

\paragraph{Sample Translation Process:} In the Table \ref{table:ex}, we compare the process of generating the target sequence using \textit{ReTA} and \textit{FReTNA} models. The examples are collected by running inference using these two models on the DeEn test set. The \textit{ReTA} generates the target tokens from scratch at every step in an autoregressive manner which leads to a high inference time. On the other hand, our \textit{FReTNA} model generates the target sequence parallelly by inserting/deleting multiple tokens at each step. We included only one example here due to space constraints; more examples can be found in the Appendix.

\section{Related Work}
\paragraph{Simultaneous Translation:}
The earlier works in streaming simultaneous translation such as  \citet{cho2016can,gu2016learning,press2018you}
lack the ability to anticipate the words with missing source context. The \textit{Wait-k} model introduced by \citet{ma2018stacl} brought in many improvements by introducing a simultaneous translation module which can be easily integrated into most of the sequence to sequence models.
\citet{arivazhagan2019monotonic} introduced MILk which is capable of learning an adaptive schedule by using hierarchical attention; hence it performs better on the latency quality trade-off.
\textit{Wait-k} and MILk are both capable of anticipating words and achieving specified latency requirements.

\paragraph{Re-translation: }
Re-translation is a simultaneous translation task in which revisions to the partial translation beyond strictly appending of tokens are permitted. Re-translation is originally investigated by \citet{niehues2016dynamic, Niehues2018}. More recently, \citet{arivazhagan2020retranslation} extends re-translation strategy by prefix augmented training and proposes a suitable evaluation framework to assess the performance of the re-translation model. They establish re-translation to be as good or better than state-of-the-art streaming systems, even when operating under constraints that allow very few revisions.

\paragraph{Non-Autoregressive Models:}
 Breaking the autoregressive constraints and monotonic (left-to-right) decoding order in classic neural sequence generation systems has been investigated. \citet{NIPS2018_8212, wang-etal-2018-semi-autoregressive} design partially parallel decoding
schemes which output multiple tokens at each step. \citet{gu2017non} propose a non-autoregressive framework which uses discrete latent variables, and it is later adopted in \citet{lee2018deterministic} as an iterative refinement process. \citet{ghazvininejad2019mask} introduces the masked language modelling objective from BERT \cite{devlin2018bert} to non-autoregressively predict and refine the translations.
\citet{welleck2019non, stern2019insertion, gu2019insertion} generate translations non-monotonically by adding words
to the left or right of previous ones or by inserting words in arbitrary order to form a sequence.
\citet{gu2019levenshtein} propose a non-autoregressive Transformer model based on Levenshtein distance to support insertions and deletions. This model achieves a better performance and decoding efficiency compared to the previous non-autoregressive models by iteratively doing simultaneous insertion and deletion of multiple tokens. 

We leverage the non-autoregressive language generation principles to build efficient re-translation systems having low inference time.

\section{Conclusion}
The existing re-translation model achieves better or comparable performance to the streaming simultaneous translation models; however, high inference time remains as a challenge. In this work, we propose a new approach for re-translation based simultaneous translation by leveraging non-autoregressive language generation. Specifically, we adopt the Levenshtein Transformer since it is inherently trained to find corrections to the existing (partial) translation. We also propose a new stability metric which is more sensitive to the flickers in the output stream. As observed from the experimental results, the proposed approach achieves comparable translation quality with a significantly less computation time compared to the previous autoregressive re-translation approaches.

\bibliography{aaai21}

\end{document}